\documentclass{egpubl}
\usepackage{VMV2023}

\usepackage[T1]{fontenc}
\usepackage{dfadobe}  
\usepackage{placeins}
\usepackage{amsmath}

\biberVersion
\BibtexOrBiblatex
\usepackage[backend=biber,bibstyle=EG,citestyle=alphabetic,backref=true]{biblatex} 
\electronicVersion
\PrintedOrElectronic

\ifpdf \usepackage[pdftex]{graphicx} \pdfcompresslevel=9
\else \usepackage[dvips]{graphicx} \fi

\usepackage{egweblnk} 

\graphicspath{images}
\usepackage{placeins}

\title[Animation of Neural Head Avatars]%
      {Video-Driven Animation of Neural Head Avatars}

\author[W. Paier, P. Hinzer, A. Hilsmann, P.Eisert]
{\parbox{\textwidth}{\centering W. Paier$^{1}$\orcid{0000-0002-0834-0566},
         P. Hinzer$^{1}$
		 A. Hilsmann$^{1}$\orcid{0000-0002-2086-0951},
		 P. Eisert$^{1,2}$\orcid{0000-0001-8378-4805},
        }
        \\
{\parbox{\textwidth}{\centering $^1$Fraunhofer Heinrich Hertz Institute, Berlin, Germany\\
         $^2$Humboldt University, Berlin, Germany
       }
}
}

\begin{document}

\teaser{
 \includegraphics[trim=0mm 143mm 80mm 0mm,clip,width=\linewidth]{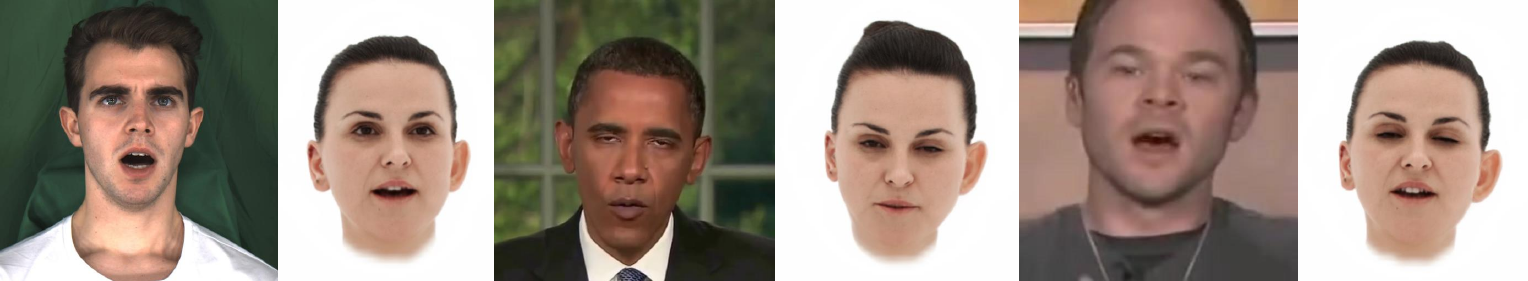}
 \centering
  \caption{Results of our video-driven animation approach trained with a single-person dataset.}
\label{fig:teaser}
}

\maketitle
\begin{abstract}

   We present a new approach for video-driven animation of high-quality neural 3D head models, addressing the challenge of person-independent animation from video input.
   Typically, high-quality generative models are learned for specific individuals from multi-view video footage, resulting in person-specific latent representations that drive the generation process.
   In order to achieve person-independent animation from video input, we introduce an LSTM-based animation network capable of translating person-independent expression features into personalized animation parameters of person-specific 3D head models.
   Our approach combines the advantages of personalized head models (high quality and realism) with the convenience of video-driven animation employing multi-person facial performance capture.
   We demonstrate the effectiveness of our approach on synthesized animations with high quality based on different source videos as well as an ablation study.


\end{abstract}  
\section{Introduction}
Analysis and synthesis of human faces play an important role in many fields such as movie productions, game development, or virtual reality (VR).
Especially, video-driven facial animation receives great attention as this technique simplifies otherwise difficult tasks such as video/photo editing, visual dubbing, or the animation of 3D human characters.
While recent advances in multi-person facial re-enactment have significantly improved animation quality, challenges persist in achieving both (photo) realism and seamless integration into 3D virtual environments, crucial for creating immersive VR experiences.

On the other hand, high-quality 3D neural head avatars are often created from captured multi-view data of a single individual, resulting in person-specific latent representations driving the generation process. This presents a significant challenge when training multi-person capable video-driven animation models. 

In this paper, we propose a new approach for multi-person capable video-driven animation of high-quality 3D neural head avatars, bridging the gap between realism and convenient animation for VR.
Our method overcomes the limitations of existing techniques by seamlessly integrating (photo) realism of personalized head avatars into multi-person video-based animation.
We employ a hybrid head representation that combines 3D mesh-based geometry, dynamic textures, and neural rendering~\cite{Paier2023}.
In order to drive our neural head model with video footage of an arbitrary person, we extract subject-independent expression features using the method of Feng et al.~\cite{Feng2021}.
Taking into account the ambiguous mapping between source expression space (person-independent) and target expression space (animated head model), we design our animation model as a recurrent neural network (LSTM) that performs not only a frame-by-frame prediction of animation parameters but considers temporal relationships as well,
resulting in more accurate animations even with unforeseen actors.
To further improve the animation quality, we augmented the extracted expression features with a learned residual, which simplifies finding a good mapping between source and target expression parameters.

The remainder of this paper is structured as follows.
The next section reviews related work before section \ref{sec:overview} presents an overview of the proposed approach.
Sections \ref{sec:headmodel} and \ref{sec:animation} describe the employed neural head model and the video-driven animation approach.
Finally, sections \ref{sec:results} and \ref{sec:conclusions} discuss the experimental results and draw a conclusion.

\section{Related Work}
\subsection*{Face Modelling}
Learning controllable models of human faces/heads has been extensively studied over the last decades.
While being comparably simple, morphable models~\cite{Blanz1999} are still one of the most popular approaches to represent facial expressions as they can be easily extended to increase the quality of facial expressions~\cite{Blanz1999, Weise2011, Garrido2013, Li2013, Cao2014}, incorporate additional attributes such as  identity, texture, and light~\cite{Vlasic2005, Thies2015} or serve as a face geometry prior for various deep-learning-based methods~\cite{Tewari2017, Kim2018, Chai2020, Mallikarjun2021, Gafni2021, Grassal2022}.
However, a significant drawback of these linear models is the lack of expressiveness to correctly represent complex areas like the oral cavity, eyes, or hair.
As a result, purely model-based approaches often employ 'hand crafted' solutions (e.g.~oral cavity) or simply ignore these areas~\cite{Garrido2013, Thies2015}, while hybrid or 2D methods tackle this problem by representing facial performances in geometry and texture space~\cite{Paier2017, Dale2011} or directly in 2D image space~\cite{Hong2022, Wang2021, Siarohin2019, Isola2016}.
In contrast, neural face models are based on deep generative architectures such as variational auto-encoder~\cite{kingma2013autoencoding} or generative adversarial networks~\cite{Goodfellow2014}, 
which can synthesize detailed 3D geometry and high-resolution face textures from a latent expression vector~\cite{Bi2021, Chandran2020, Li2020, paier2020, Lombardi2018}.
More recently, Ma et al.~\cite{Ma2021} proposed a codec avatar that supports inference and rendering of neural head models even on mobile devices, whereas Grassal et al.~\cite{Grassal2022} present an approach for creating a personalized neural head avatar even from monocular video.
Apart from traditional mesh and texture-based models, several novel scene representations~\cite{Mildenhall2019, Sitzmann2019, Lombardi2019b, Pumarola2020, Mildenhall2020, Tewari2020STAR, Yu2021, Reiser2021, Hedman2021, Gafni2021} have been presented in the recent years that allow for creating truly photo-realistic renderings of humans.
For example, Thies et al.~\cite{Thies2019} propose a deferred neural rendering approach to create photo-realistic renderings of 3D computer graphic models.
Kim et al.~\cite{Kim2018} and Prokudin et al.~\cite{Prokudin2020} combine neural rendering with parametric models of humans and human heads, which allows for photo-realistic rendering of animatable CG models.
Volumetric representations~\cite{Mildenhall2019, Sitzmann2019, Lombardi2019b, Pumarola2020, Mildenhall2020, Tewari2020STAR} often capture the 3D structure and appearance of an object/scene with a non-linear function that depends on the 3D position in space as well as viewing direction and predicts color as well as volume density.
This allows for representing fine structures such as hair fibers, objects with complex reflective properties such as glass and metal but also dynamic effects like smoke.
A big drawback, however, is the high computational complexity \cite{Yu2021, Reiser2021, Hedman2021, Gafni2021}.
While M\"uller et al.~\cite{Mueller2022} propose a new approach that supports very fast training as well as real-time rendering, they still lack semantic control, which is essential for animation.

\subsection*{Video-driven Facial Animation}
In recent years, many methods for multi-person video-driven facial animation have been published, which can be categorized into 2D and 3D approaches.

The main advantage of 2D approaches~\cite{Zeng2023, Drobyshev2022, Wang2021, Ha2020, Zeng2020, Siarohin2019, Geng2018} is their simplicity of use as a single image of the target person can be animated with a driving-video of another person requiring no additional preprocessing.
A big disadvantage, however, is the lack of 3D information, which results in unrealistic distortions and obvious rendering artifacts as soon as the head pose changes considerably.
Several approaches have been proposed to overcome this problem.
For example, Wang et al.~\cite{Wang2021} predict explicit features for appearance, expression, head-pose, as well as canonical 3D key points, whereas
Hong et al.~\cite{Hong2022} introduce a depth-aware generative adversarial network (GAN) that predicts depth values for each face pixel.
While both 2D methods can improve the visual quality of resulting videos, the lack of a consistent underlying 3D model still degrades the rendering quality under strong 3D rotations.
More importantly, these methods cannot be used to animate faces for 3D applications such as games or virtual reality.

3D-aware methods~\cite{Filntisis2022, Danecek2021, Feng2021, Thies2019, Kim2018} usually fit an existing 3D morphable face model to each frame of a video by predicting the corresponding model parameters (i.e. identity, expression, head pose, etc.) with a convolutional neural network.
This captures the 3D facial performance of a person from a monocular video and allows for transferring the captured facial expression to a target person in a different video (e.g. by rendering the animated face model with adapted expression parameters).
While this approach works well in general, the underlying linear face models often do not capture/reproduce facial expressions accurately enough and/or they cannot be easily integrated into real-time 3D environments.

\subsection*{Contribution}
\begin{figure*}[htb]
\centering
\includegraphics[trim=0mm 110mm 0mm 0mm,clip,width=\textwidth]{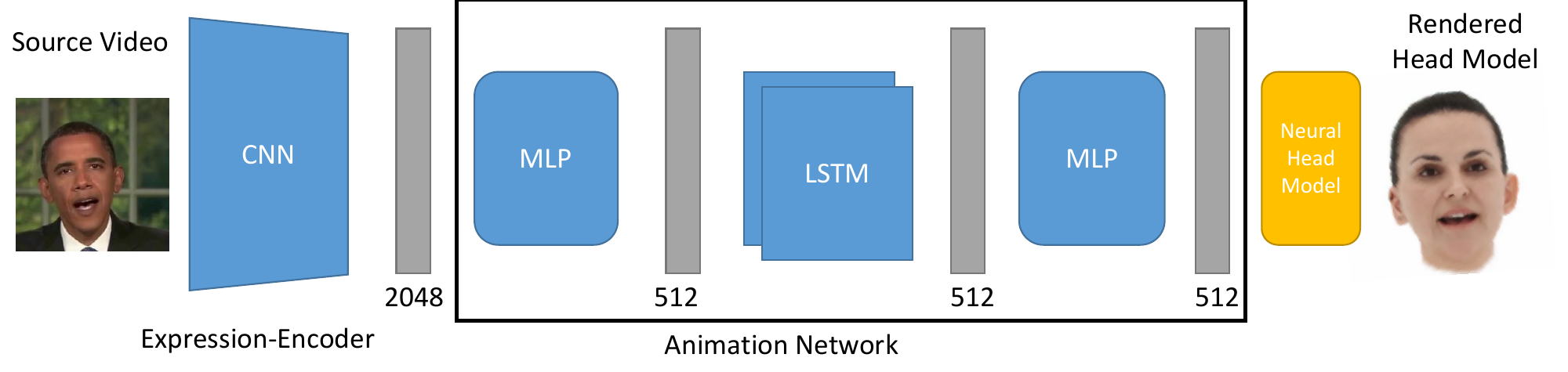}
\protect\caption{High-level architecture of the proposed animation network.}
\label{fig:overview}
\end{figure*}
In this paper, we present a new approach for real-time video-driven animation of a personalized 3D neural head model.
Unlike previous methods in multi-person video-driven animation, our approach surpasses the limitations associated with linear morphable face models that frequently exhibit insufficient accuracy in capturing and reproducing intricate facial expressions. 

Instead, we employ a personalized high-quality neural head avatar that allows for photo-realistic rendering and convenient integration in 3D scenes. 
In order to bridge the gap between the driving video and our neural head model, we extract person-independent expression features that allow for transferring the facial expression from the source video to our neural head avatar using a recurrent neural network (LSTM).
Our contribution thereby enhances the fidelity and realism of video-driven facial animations and therefore allows effortless integration into immersive experiences in virtual environments.

\section{System Overview}
\label{sec:overview}
In the first step, we create a personalized neural head model that represents 3D geometry, head motion, facial expression, and appearance of the captured person, Sec.~\ref{sec:headmodel}.
This model is learned from video data of an actor captured with a multi-view camera rig and can be driven with a latent expression vector.
After training the VAE-based head model, the captured facial performance can be represented by a sequence of low-dimensional latent parameter vectors.
In the second step, we compute person-independent expression features for each captured video frame using the approach of Feng et al.~\cite{Feng2021}.
Hence, the complex task of video-driven animation is reduced to transforming a sequence of person-independent expression features into a sequence of animation parameters for our neural head model using a recurrent neural network, Sec.~\ref{sec:animation}.
Finally, to increase realism, we employ a neural rendering model that refines rasterization-based images of our head model, which allows for synthesizing photo-realistic videos of the animated head model.
\FloatBarrier

\section{Hybrid Head Representation}
\label{sec:headmodel}

Our approach is based on a photo-realistic animatable 3D head model~\cite{Paier2023} that is computed from multi-view video footage to ensure that it
perfectly resembles the appearance of the captured person.
The model-building process consists of three stages: first, a statistical head model is employed to recover pose and approximate head geometry for each captured frame based on automatically detected landmarks~\cite{Kazemi2014} and optical flow.
In the second step, dynamic head textures are extracted in addition to the approximate geometry to reproduce fine details, small motions, and complex deformations (e.g.~in the oral cavity).
After geometry recovery and texture extraction, each captured frame is represented by rigid motion parameters $\mathbf{T}$, blend-shape weights $\mathbf{b}$, and an RGB image as texture.
Based on this data, we train a deep generative face model (VAE) that reconstructs blend-shape weights $\mathbf{b}$ as well as face textures from a low-dimensional expression vector $\mathbf{z}$, thereby enabling natural, plausible, and realistic facial animation.
An adversarial training strategy based on a patch-based discriminator network~\cite{Isola2016} helps to improve the texture reconstruction quality.
After the model creation, all captured multi-view sequences are represented with a single parameter vector consisting of 3D head pose $\mathbf{T}$ as well as expression vectors $\mathbf{z}$.

\begin{figure}[htb]
\centering
\includegraphics[trim=0mm 0mm 88mm 0mm,clip,width=\columnwidth]{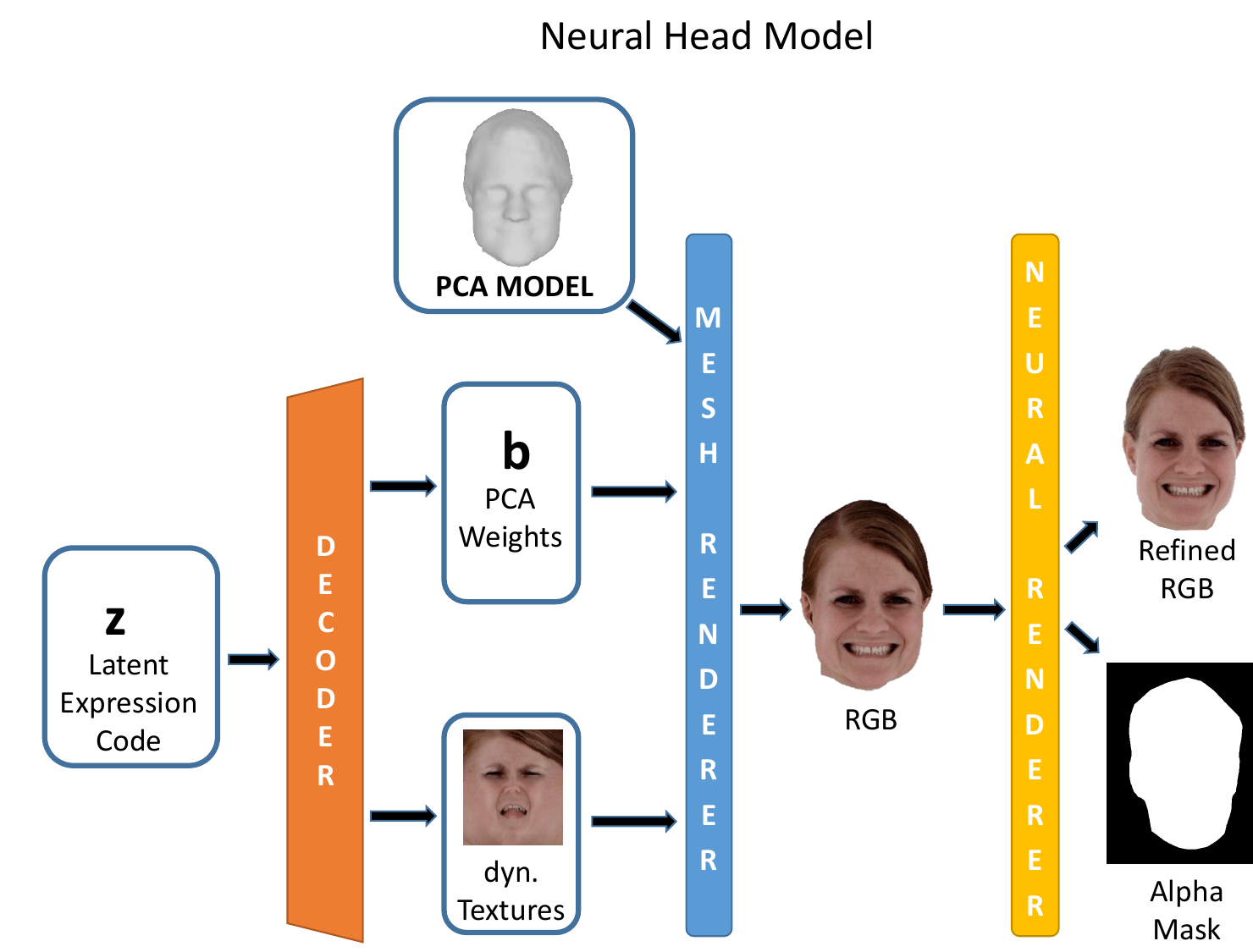}
\protect\caption{High-level architecture of the employed neural head model.}
\label{fig:neural-head}
\end{figure}

\begin{figure*}[t!]
\centering
\includegraphics[trim=0mm 35mm 0mm 0mm,clip,width=0.9\textwidth]{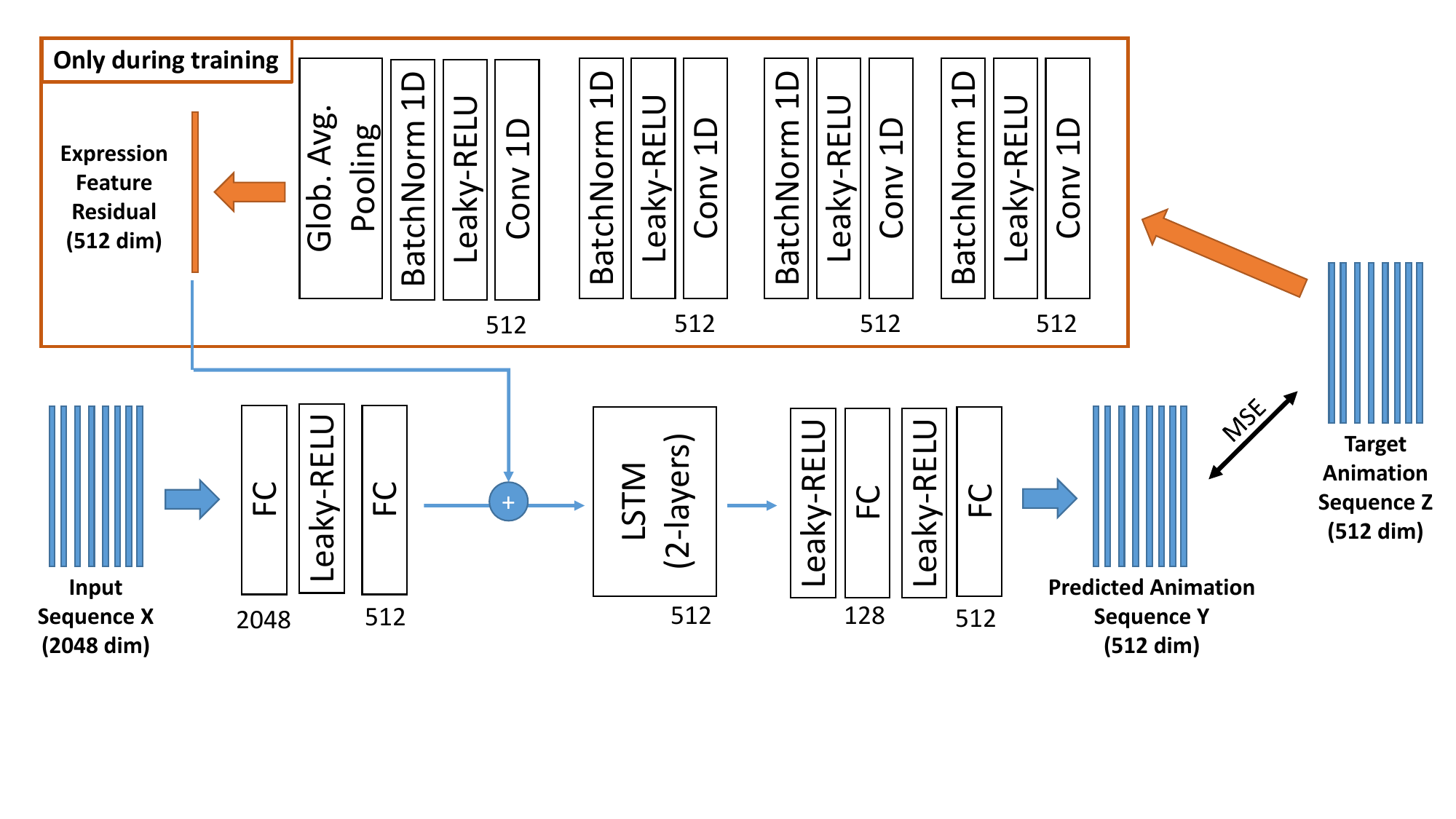}
\protect\caption{Architecture of the animation network.}
\label{fig:anim-arch}
\end{figure*}

\begin{figure}[h]
\includegraphics[trim=0mm 92mm 130mm 0mm,clip,width=\columnwidth]{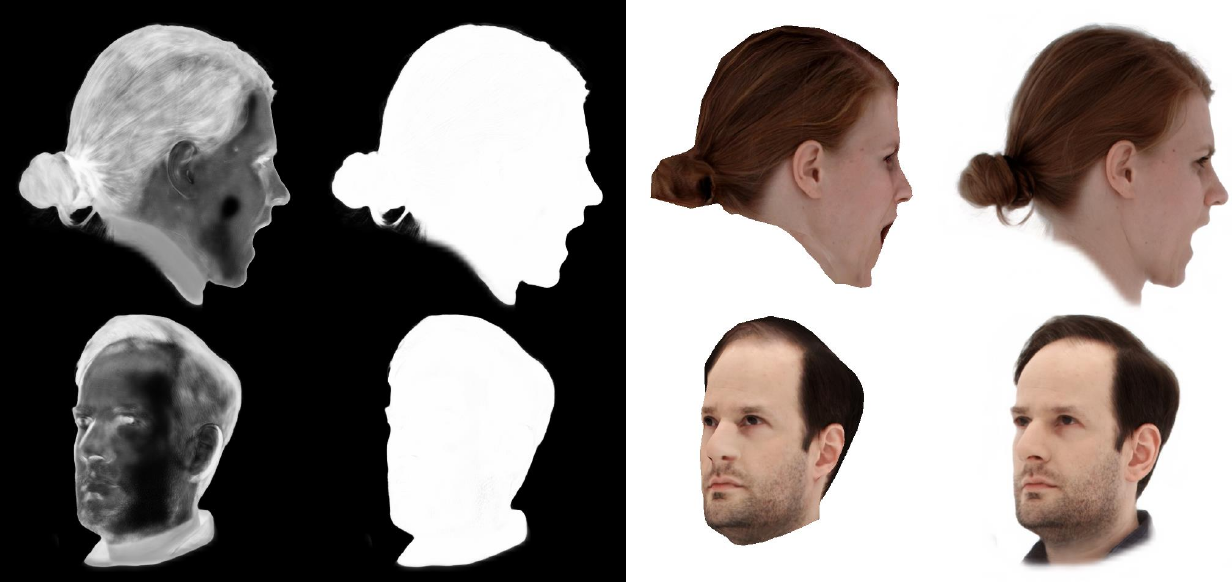}
\protect\caption{This figure illustrates the quality of the employed hybrid head model. The leftmost column shows the refinement weights $\beta$ followed by the foreground mask $\mathcal{F}$, the initial mesh-based rendering, and the final output $\mathcal{I}_{out}$. High intensity in the refinement mask indicates strong corrections (e.g.~neck, hair, sometimes mouth), while low intensity indicates that the mesh-based rendering provides already correct pixel colors.}
\label{fig:render-input-output}
\end{figure}

While the convenient mesh-plus-texture representation allows for correctly reconstructing the facial performance as well as most of the appearance, important details (e.g.~silhouette, oral cavity, hair, etc.) cannot be reproduced, see figure \ref{fig:render-input-output}.
Therefore, a self-supervised rendering approach based on pixel-to-pixel translation is employed.
The render network receives the mesh-based rendering as input and predicts a refined head image as well as weight masks that help to separate foreground from background.
This simplifies the training process (i.e. no need to pre-compute foreground masks) and provides a means for integrating the rendered head model with new backgrounds or into 3D scenes.

The image formation model (\ref{eq:imageformation}) is a convex combination of the mesh-based rendering $\mathcal{I}_{orig}$, a corrective image $\mathcal{I}_{corr}$ and the static background $\mathcal{I}_{backg}$, where each image contributes according to spatially varying weight maps $\alpha$, $\beta$, and $\gamma$.

\begin{equation}
\begin{split}
    \mathcal{I}_{out} = \alpha\ \mathcal{I}_{orig} + \beta\ \mathcal{I}_{corr} + \gamma\ \mathcal{I}_{backg} \\
    \alpha + \beta + \gamma = 1
\end{split}
\label{eq:imageformation}
\end{equation}

The real-time re-rendering network is based on a U-Net architecture \cite{Ronneberger2015} where the input tensor contains the RGB colors of the mesh-based rendering and the output tensor consists of six channels: RGB color plus three channels that contain the weight maps $\alpha$, $\beta$, and $\gamma$.
Training the render model requires only the captured frame, the mesh-based rendering $\mathcal{I}_{orig}$ of the textured head model, an empty background frame (i.e. clean plate) $\mathcal{I}_{backg}$ while an alpha-mask for foreground/background segmentation is learned automatically in an unsupervised manner.
For more details please refer to \cite{Paier2023}.
\section{Video-Based Neural Animation}
\label{sec:animation}
After the creation of the neural head model, each frame of the captured multi-view video footage is encoded into a low dimensional latent vector $\mathbf{z}$.
This vector represents the facial expression and head pose of the actor captured in the footage.
However, as the training database only consists of visual information of a single individual, training a multi-person animation approach directly poses a challenge.
To overcome this limitation, we extract person-independent expression features $\mathbf{x}$ from the captured video footage using the method of Feng et al.~\cite{Feng2021}, which was trained with a large number of individuals showing different facial expressions and head poses.
This way, each frame of our multi-view data is labeled with a person-independent expression information $\mathbf{x}$ as well as a latent expression vector $\mathbf{z}$ that drives our neural head model.
In order to animate the neural head avatar, it is essential to predict the corresponding animation parameter $\mathbf{z}$ from each expression feature $\mathbf{x}$.
Due to the inherent ambiguity of this mapping, we employ a recurrent architecture (LSTM) that performs not only a frame-wise mapping but also captures temporal relationships between sequences of input features and animation parameters, figure \ref{fig:anim-arch}.

\begin{figure*}[t!]
\centering
\includegraphics[trim=0mm 65mm 0mm 45mm,clip,width=0.8\textwidth]{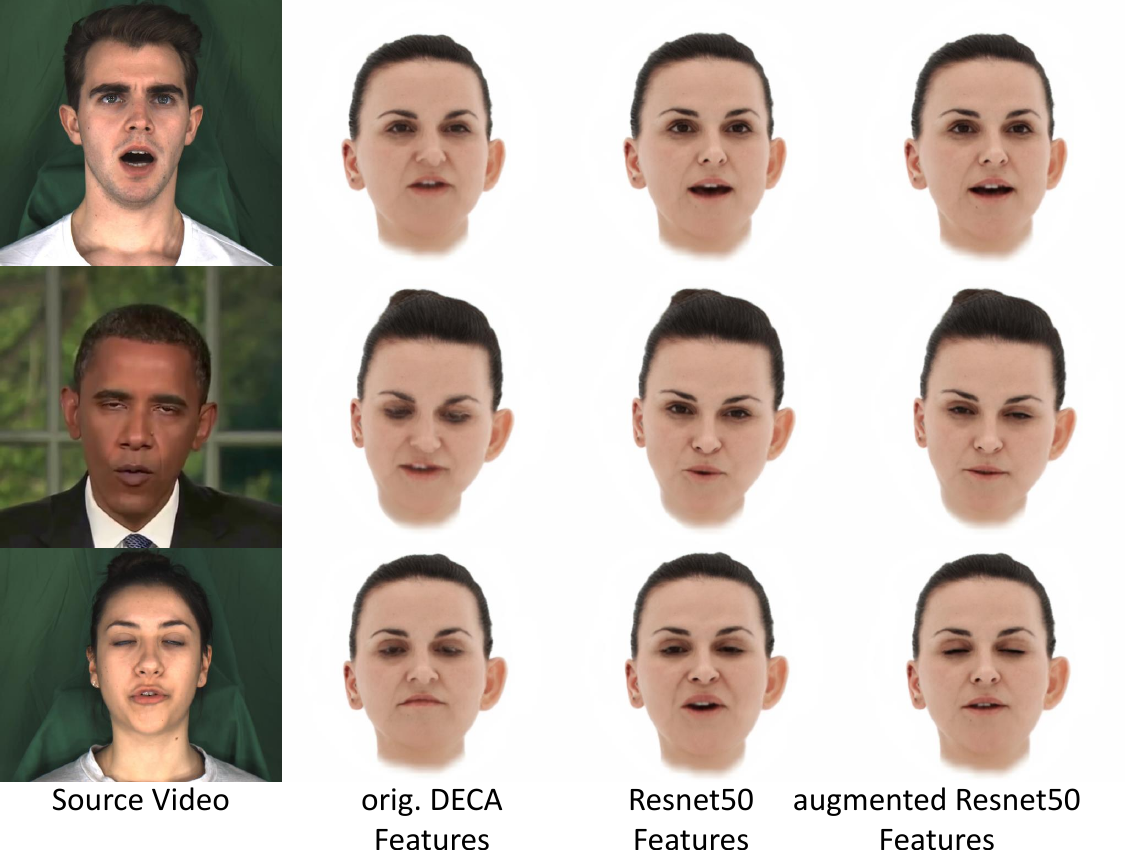}
\protect\caption{Visual comparison of the facial animation based on different input features.}
\label{fig:ablation}
\end{figure*}

However, since the predicted expression weights and jaw angles of Feng et al.~\cite{Feng2021} are optimized for the FLAME face model~\cite{Li2017}, we use the 2048-dimensional output feature vector of an earlier layer (output of the ResNet50).
During our experiments, we found that computing a small expression residual feature (during training) with a CNN from the target animation parameters helps to further improve the animation quality.
This CNN consists of four 1D convolutions (with kernel size 5, zero-padding of 2) followed by a Leaky-ReLU and BatchNorm.
In the end, an average-pooling layer outputs the mean expression residual over all 8 frames, which is then injected into the animation network, see \ref{fig:anim-arch}.
Our animation network is trained batch-wise (32) with short sequences (8 frames) of $\mathbf{x}$ transformed into equally long sequences of $\mathbf{y}$ minimizing the mean squared error (MSE) between $\mathbf{y}$ and the target expression $\mathbf{z}$.
All Leaky-ReLU layers have a leakiness of $1.0e-2$.
We train the network for 15000 iterations using the Adam optimizer with a learning rate of $1.0e-4$ and exponential learning rate scheduling with $\gamma=0.96$.

\section{Experimental Results and Discussion}
\label{sec:results}
This section presents our evaluation, and visual results generated with the proposed method as still images, while an accompanying video demonstrating dynamic effects can be found in the supplementary material.
For our experiments, we captured an actress with a synchronized and calibrated multi-view camera rig consisting of three cameras (frontal, diagonally left/right) at eye level.
We captured different facial expressions as well as speech (single words and monologues in English).
The actress was not restricted in terms of the presented emotion.
The effective capture resolution for the head is approximately 520x360 pixels.
All pre-processing steps, network training, and experiments have been carried out on a regular desktop computer with 64GB Ram, 2.6 GHz CPU (14 cores with hyper-threading), and one GeForce RTX3090 graphics card.
The captured data was split into four sequences with a total length of approximately 3 minutes for training and one test sequence with a total duration of approximately 30 seconds.
We evaluated the proposed network architecture with different input features: the original DECA expression features (blend shape weights and jaw angles), earlier expression features produced by the Resnet50, and Resnet50 features augmented with an auxiliary feature vector that helps to disambiguate the mapping between input expression feature and target animation parameter. For inference, we use a zero residual feature vector for animation.

During development, we visually compared an MLP-based, CNN-based, and LSTM-based architecture and found that the LSTM-based performs best.
Moreover, we use rather short input sequences of 8 frames as we found that more temporal context does not yield better animation results anymore.

\begin{figure*}[h!]
\centering
\includegraphics[trim=0mm 0mm 0mm 0mm,clip,width=0.9\textwidth]{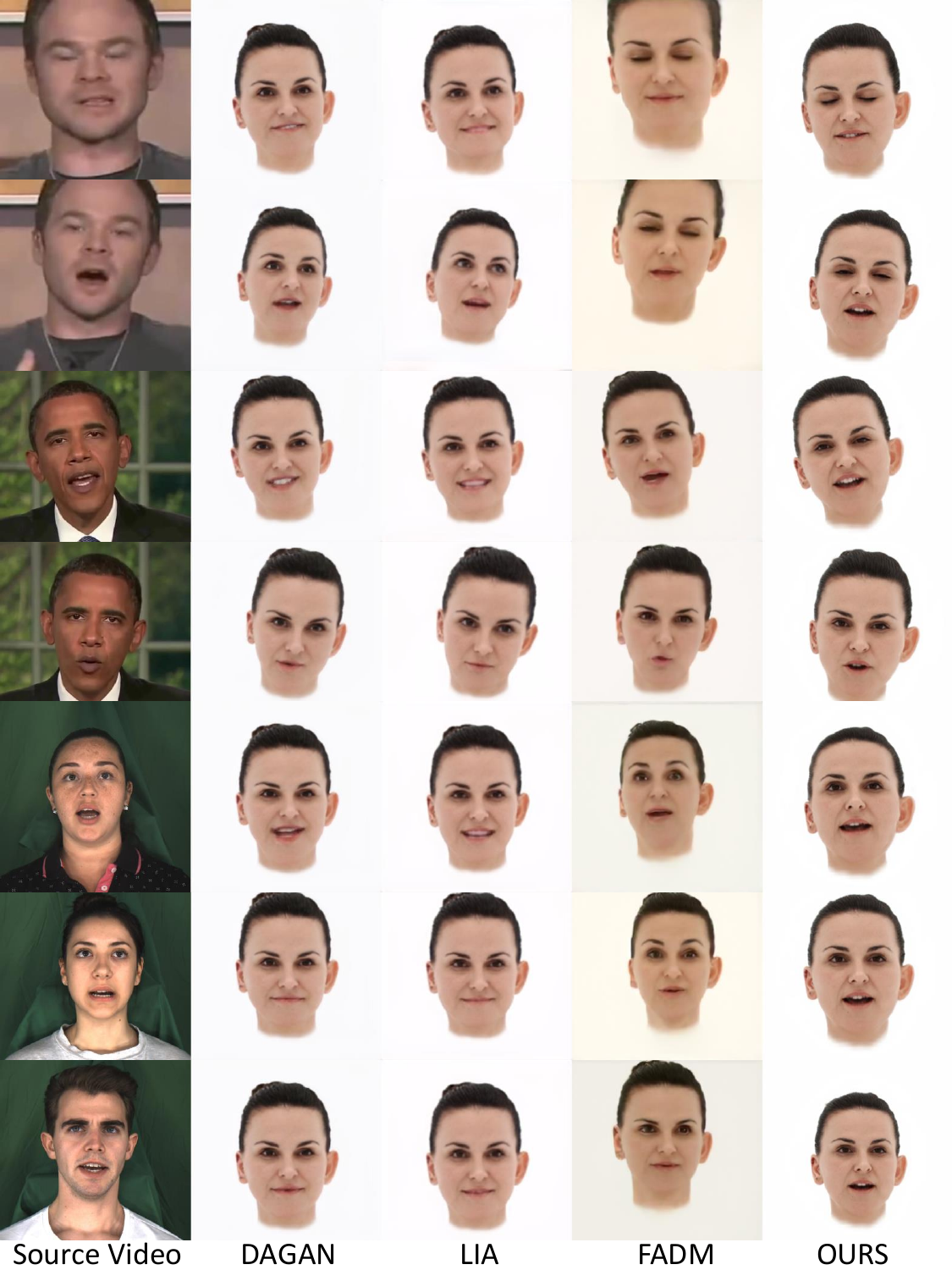}
\protect\caption{Comparison of the proposed animation approach with 3 recent methods for multi-person video-driven face animation.}
\label{fig:comparison}
\end{figure*}

Figure \ref{fig:ablation} illustrates the differences in the resulting animation based on the tested input features.
In our experiments, we found that the generated talking head videos based on the Resnet50 features appear to be more realistic and included fewer artifacts.
Especially, the augmented Resnet50 features yield livelier and more natural animations, which can be seen in the supplemental video.
We evaluate our animation method against three recent approaches (DAGAN~\cite{Hong2022}, LIA~\cite{Wang2022}, FADM~\cite{Zeng2023}) for the generation of photo-realistic talking heads based on the driving video of an arbitrary person.
Figure \ref{fig:comparison} illustrates the rendering and animation quality of all approaches based on video samples taken from the VoxCeleb2~\cite{Chung2018} dataset, MEAD~\cite{Wang2020} dataset, and Obama's weekly address footage~\cite{Suwajanakorn2017}.
The main advantage of our approach is that a high-quality neural head model can be connected with a multi-person capable video-driven animation approach, which results in the higher visual quality of the synthesized videos, more visible details, fewer rendering artifacts, and more natural animations.
Our intuition on why the residual features improve animation during inference is that they reduce the likelihood that the network learns spurious correlations between input expression features and target animation parameters.
Expression differences that can neither be explained by the original Resnet50 features nor by the temporal context can be represented by the residual features, which are computed from the target animation parameters.
Additionally, providing only an average residual feature per training sequence prevents the network from relying too much on the these artificial features.

There are also limitations: in order to achieve high visual quality a personalized neural head model is created, however this requires retraining if a new virtual character has to be integrated. 
Currently, the residual features are only used during training but not for inference. However, with a suitable generative model, they could enable fine-tuning of the generated animations also during inference as they capture further expression details that cannot be explained by the original input features.

\section{Conclusions}
\label{sec:conclusions}
We present a new method for the animation of 3D neural head models.
Our method extracts person-independent expression features from monocular video and translates them successfully into realistic animation parameters for our neural head model.
This allows for animating high-quality 3D head avatars by arbitrary actors even though the model is generated only from captured data of a single person.
For more robust training, we augment the extracted expression features, which helps to disambiguate the mapping between source expression features and target animation space.
We show that our neural head model can be successfully animated from arbitrary persons and compare our approach against recent methods for video-driven facial re-enactment demonstrating the high quality of our animation results.
\section*{Acknowledgments}
This work has partly been funded by the European Union under grant agreement No 101070672 (SPIRIT),
by the German Federal Ministry of Education and Research (Voluprof, grant no.~16SV8705),
by the German Federal Ministry for Economic Affairs and Climate Action (ToHyVe, grant no.~01MT22002A)
as well as the Fraunhofer Society in the Max Planck-Fraunhofer collaboration project NeuroHum.
\printbibliography

\end{document}